\documentclass{article}

\usepackage{arxiv}
\usepackage[utf8]{inputenc} % allow utf-8 input
\usepackage[T1]{fontenc}    % use 8-bit T1 fonts
\usepackage{hyperref}       % hyperlinks
\usepackage{url}            % simple URL typesetting
\usepackage{booktabs}       % professional-quality tables
\usepackage{amsfonts}       % blackboard math symbols
\usepackage{nicefrac}       % compact symbols for 1/2, etc.
\usepackage{microtype}      % microtypography
\usepackage{multirow}
\usepackage{graphicx}
\usepackage{caption}
\usepackage{subcaption}
\usepackage{threeparttable}
\usepackage{moresize}
\usepackage{siunitx}

\newcommand{\Description}[2][]{}
\newenvironment{acks}{\section*{Acknowledgments}}

\title{Towards Hardware-Agnostic Gaze-Trackers}

\author{
  Jatin ~Sharma \\
  Microsoft Research\\
  Redmond, WA 98052 \\
  \texttt{jashar@microsoft.com} \\
  \And
  Jon ~Campbell \\
  Microsoft Research\\
  Redmond, WA 98052 \\
  \texttt{joncamp@microsoft.com} \\
  \And
  Pete ~Ansell \\
  Microsoft Research\\
  Redmond, WA 98052 \\
  \texttt{petea@microsoft.com} \\
  \And
  Jay ~Beavers \\
  Microsoft Research\\
  Redmond, WA 98052 \\
  \texttt{jbeavers@microsoft.com} \\
  \And
  Christopher O'Dowd \\
  Microsoft Research\\
  Redmond, WA 98052 \\
  \texttt{chriod@microsoft.com} \\
}

\begin{document}

%%%%%%%%%%%%%%%%%%%%%%%%%%%%%%%%%%%%%%%%%%%%%%%%%%%%%%%%%%%%%%%%%%%%%%%%%%%%%%%%%%%%%%%%%%%%%%%

%% This command processes the author and affiliation and title
%% information and builds the first part of the formatted document.
\maketitle

\begin{abstract}
Gaze-tracking is a novel way of interacting with computers which allows new scenarios, such as enabling people with motor-neuron disabilities to control their computers or doctors to interact with patient information without touching screen or keyboard. Further, there are emerging applications of gaze-tracking in interactive gaming, user experience research, human attention analysis and behavioral studies. Accurate estimation of the gaze may involve accounting for head-pose, head-position, eye rotation, distance from the object as well as operating conditions such as illumination, occlusion, background noise and various biological aspects of the user. Commercially available gaze-trackers utilize specialized sensor assemblies that usually consist of an infrared light source and camera. There are several challenges in the universal proliferation of gaze-tracking as accessibility technologies, specifically its affordability, reliability, and ease-of-use. In this paper, we try to address these challenges through the development of a hardware-agnostic gaze-tracker. We present a deep neural network architecture as an appearance-based method for constrained gaze-tracking that utilizes facial imagery captured on an ordinary RGB camera ubiquitous in all modern computing devices. Our system achieved an error of 1.8073cm on GazeCapture dataset without any calibration or device specific fine-tuning. This research shows promise that one day soon any computer, tablet, or phone will be controllable using just your eyes due to the prediction capabilities of deep neutral networks.
\end{abstract}

% keywords can be removed

\section{Introduction}
\label{sec:introduction}

\subsection{Motivation}
\label{ssec:motivation}
Gaze-tracking is an increasingly popular field of computational research due to its emerging applications in interactive gaming, user experience research, human attention analysis, behavioral studies, and accessibility technologies. Use of gaze estimation as Accessibility Technology can help people living with various kinds of Motor Neuron Diseases (MNDs) and disorders such as Amyotrophic Lateral Sclerosis (ALS) and Cerebral Palsy (CP), and injuries such as Spinal Cord Injury (SCI) and Traumatic Brain Injury (TBI). Eye-gaze based computers enable such users to exert control over their environment and communicate with others. However, estimating where a user is looking, is a complex computational task that may require accounting for user's head-pose, head-positioning, eye rotation, and the distance from the object. These variables are measured with respect to the observer’s frame of reference. The observer is generally an assembly of infrared light projector and a high-resolution infrared camera. Variations in illumination, background noise, various biological aspects of the user as well as the presence of glasses, face coverings and assistive medical equipment may also impact the accuracy. Further, the variations in the system configurations such as optical properties of the sensors and quality of imaging bring in additional set of variability. Some devices work by comparing several images taken under different lighting and using the physiology of eyes to locate them and to estimate their pose. A red-eye effect indicates where the eye pupils are and a glint on the surface of the cornea allows it to estimate its orientation. Such devices may not track face or head and can rely on device or user specific calibration. This leads to development of specialized hardware that are available through limited suppliers and require significant effort in initial setup -- raising challenges in affordability, reliability, and ease-of-use.

\subsection{Contributions}
\label{ssec:contribution}
Our work makes the following contributions. First, we present the theoretical background of gaze-tracking and discuss related work in this area. Second, we reproduce the state-of-the-art iTracker \cite{Krafka16} system as a baseline to benchmark our experiments. Third, we propose a series of enhancements to develop a novel deep learning pipeline that efficiently performs gaze estimation in a hardware-agnostic fashion. And fourth, we relax the uniform data distribution constraint in the GazeCapture dataset by allowing subjects who did not look at the full set of points for evaluation and explore its impact on performance. We do not employ any calibration or device-specific fine-tuning during evaluation. The remainder of this paper is structured as follows. In Section \ref{sec:related_work}, we present a survey of the existing work in the field and discuss the state-of-the-art. In Section \ref{sec:experimental_setup}, we go through the details of our experimental setup, dataset and baseline and then conduct a series of incremental experiments in Section \ref{sec:experiments} to enhance the performance of the underlying deep neural network architecture. We demonstrate how some of the pre-processing steps and architecture changes greatly assist the model to learn faster and become robust to variations. In Section \ref{sec:results}, we present our experimental results and efforts towards explainability. We conclude in Section \ref{sec:conclusion} with our learning and potential future work.

\section{Theoretical Background \& Related Work}
\label{sec:related_work}

Gaze-tracking involves estimating a person's gaze indicators in a three-dimensional space. Popular gaze indicators include smooth pursuit, fixation \cite{Pollatsek09}, saccades \cite{Wong14}, gaze direction \cite{Kellnhofer19}, and gaze point \cite{Krafka16}. Gaze-tracking methods can be categorized into -- \textit{Model-based methods} that fit a geometric eye model to the eye image, or \textit{Appearance-based methods} that directly regress from eye images to gaze indicators using machine learning. Traditionally, model-based methods have dominated in delivering high accuracy, however, the evolution of deep learning appearance-based methods are rapidly bridging this gap. Deep learning eliminates the need for hand-crafted features and opens up a possibility for these models to explore and exploit features that were neglected by the previous methods. \textit{Unconstrained gaze applications} (e.g. surveillance, customer attention monitoring, player analysis and sports strategy) have a wide range of operating conditions (e.g. illumination, head and eye orientations). They generally measure coarse gaze indicators (e.g. gaze direction, fixations) in arbitrary natural scenes \cite{Kellnhofer19} \cite{Zhang15}. On the other hand, \textit{Constrained gaze applications} (e.g. gaming, augmentative \& alternative communication, user experience research) measure precise gaze indicators (e.g. gaze point, scan path) in low variability settings \cite{Krafka16}.

% they were superseded by technologies that were more accurate and less invasive.
Early gaze-tracking methods utilized electro-mechanical devices (e.g. blunt needle \cite{Lamare93}, stylus and drum \cite{Delabarre98}, search coils \cite{Robinson63}) in or around the eye connected to record eye movements but suffered from mechanical imperfections limiting the precision and invasiveness which discouraged general use. Erdmann \& Dodge \cite{Erdmann98} were first to pass a focused light-beam through the cornea and record it on a moveable photographic plate using an assembly of lenses. Diffractive distortion, alignment of moving parts, and restraining the participant’s head limited the efficacy of such methods. A much evolved form of this technique -- Pupil Center Corneal Reflection (PCCR) -- is used in modern gaze-trackers where the pupil is illuminated with near-infrared or infrared light and corneal reflection is captured on an infrared camera to measure eye rotation against the pupil center and deduce various gaze indicators. Please refer to \cite{Hansen10} for an extensive survey on gaze-estimation methods.

Most commercial gaze-trackers require specialized hardware setups that are available through limited suppliers and demand significant effort in initial setup -- causing challenges in affordability, reliability, and ease of use. RGB cameras are ubiquitous in all modern computing devices and combined with the recent advances in deep learning could address these challenges. This approach requires building an efficient deep neural network architecture and collecting a large amount of data to train it. Zhang et al \cite{Zhang15} created MPIIGaze dataset containing over 213 thousand images and corresponding ground-truth gaze positions from 15 laptop users over several months. They proposed GazeNet -- the first appearance-based unconstrained gaze estimation method -- that uses HOG based face landmarks to fit a standard 3D facial shape model and then use the estimated head pose for perspective warping to normalize the eye images. It reduced the mean error from \ang{13.9} to \ang{10.8}. GazeNet revealed that a good image resolution and existence of eye regions in the input images improves the precision whereas changing lighting conditions, facial appearance variations and target gaze range raise challenges in an accurate gaze estimation. Wood et al \cite{Wood16} demonstrated an effective approach to large scale data creation for difficult in-the-wild scenarios and extreme gaze angles. They combined generative 3D modeling and real-time rendering to artificially synthesize large amounts of training data with variable eye region. Their rendering technique utilized high-resolution face scans, real-time approximations for complex eyeball materials, and anatomically inspired procedural geometry methods for eye-lid animation. Krafka et al \cite{Krafka16} introduced GazeCapture, the first large-scale dataset for gaze-tracking which contained around 2.4 million frames from 1474 participants. They also introduced iTracker, a convolutional neural network to perform gaze-tracking on Apple devices using an RGB camera. The authors observed that using face, both eyes and overall face-grid as inputs provide sufficient information about head-pose, head-position and eye rotation to the network for making a reasonable gaze estimation. They used AlexNet \cite{Krizhevsky12} as the backbone network for each of the four input images. Their system achieved a prediction error of 1.86cm and 2.81cm on smartphones and tablets respectively without calibration when data augmentation was only applied to the training split. For unconstrained 3D gaze estimation Kellnhofer et al \cite{Kellnhofer19} contributed the largest publicly available dataset consisting of 172 thousand images from 238 subjects with a wide range of head poses and distances in both indoor and outdoor settings. They proposed the Gaze360 model architecture which utilizes a 7-frame sliding window centered at the target frame as an input to backbone ResNet-18 \cite{He16} network and subsequently to a bidirectional LSTM. Their system achieved mean angular error of \ang{13.5}.

%%%%%%%%%%%% Figure: Usecase and GazeCapture Dataset Characterstics %%%%%%%%%%
\begin{figure}[]
\centering
    \begin{subfigure}[t]{0.245\linewidth}
        \includegraphics[width=1\linewidth]{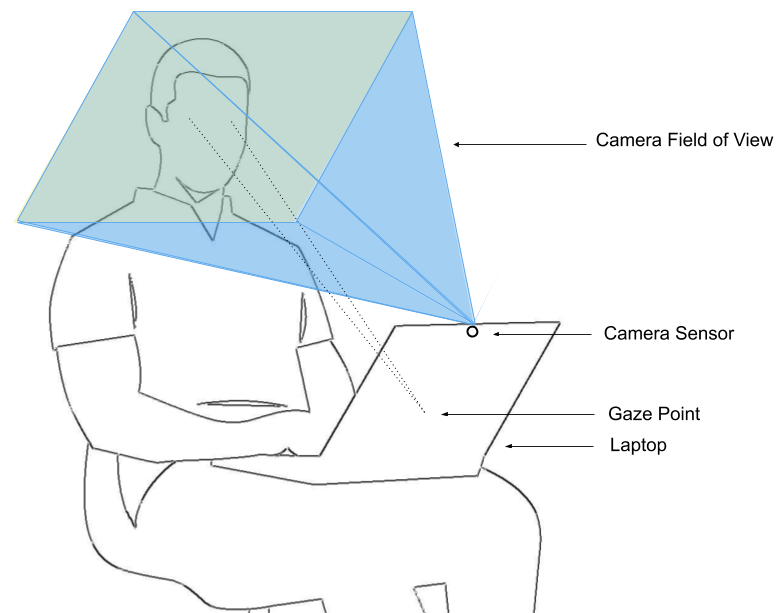}
        \caption{}\label{fig:usecase}
        \Description[A person working on a laptop]{A person working on a laptop while the in-built camera tracks person's face}
    \end{subfigure}
    % leaving a space will take the subfigure to next line
    \begin{subfigure}[t]{0.245\linewidth}
        \includegraphics[trim=0.5cm 0.2cm 1.4cm 0.9cm,clip,width=1\linewidth]{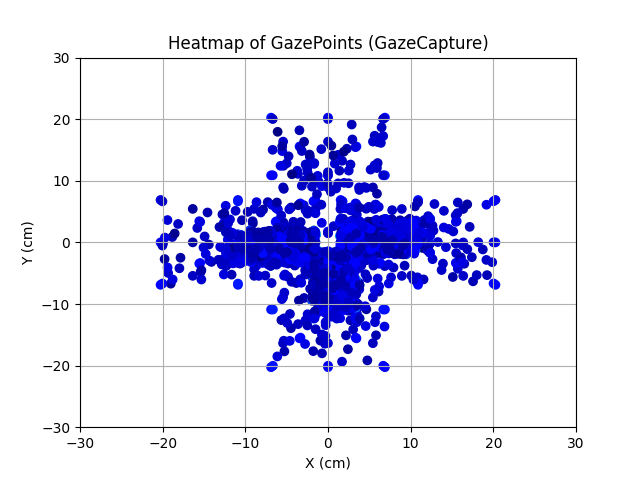}
        \caption{}\label{fig:plot_gaze_point_heatmap_gc1}
        \Description[GazeCapture data distribution]{Gaze Point distribution in GazeCapture dataset which creates a plus-sign like shape with most ground-truth points between -20cm to +20cm about both X and Y axes.}
    \end{subfigure}
    \begin{subfigure}[t]{0.245\linewidth}
        \includegraphics[trim=0.5cm 0.2cm 1.4cm 0.9cm,clip,width=1\linewidth]{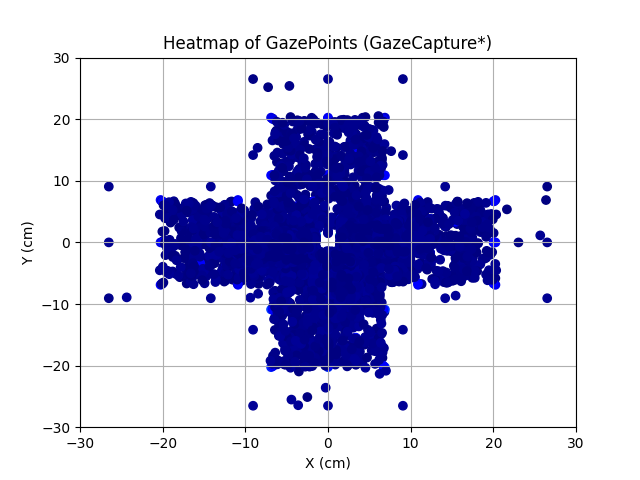}
        \caption{}\label{fig:plot_gaze_point_heatmap_gc2}
        \Description[GazeCapture* data distribution]{Gaze Point distribution in GazeCapture* dataset which creates a plus-sign like shape with most ground-truth points between -20cm to +20cm about both X and Y axes.}
    \end{subfigure}
    % leaving a space will take the subfigure to next line
    \begin{subfigure}[t]{0.245\linewidth}
        \includegraphics[trim=0.5cm 0.2cm 1.4cm 0.9cm,clip,width=1\linewidth]{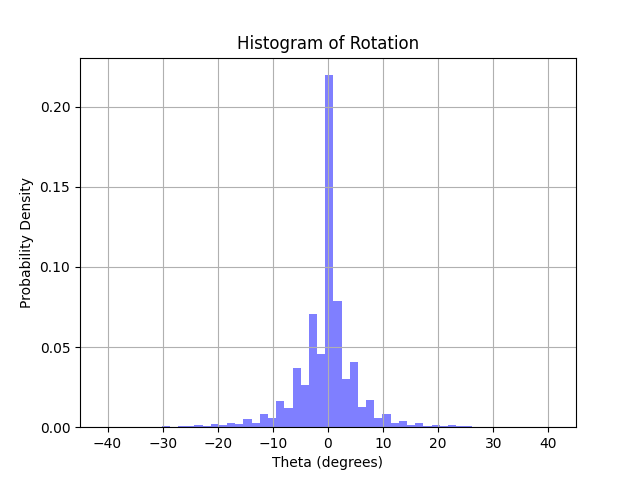}
        % \caption{Histogram of head-rotation angles}\label{fig:plot_rotation_histogram}
        \caption{}\label{fig:plot_rotation_histogram}
        \Description[A bell-like histogram]{A bell-like histogram of head rotation angles centered at \ang{0} with most measurements between \ang{-20} and \ang{+20}}
    \end{subfigure}
    % leaving a space will take the subfigure to next line
    \captionsetup{justification=centering,margin=0cm}
    \caption{(a) Gaze-tracking scenario on portable computing devices, (b)-(d) Dataset characteristics}
    % \caption{(a) A typical constrained eye-tracking scenario on mobile computing devices. (b) \& (c) Distribution of gaze points in prediction space in GazeCapture and GazeCapture* dataset}
    \label{fig:introduction}
    \Description[Use case scenario and dataset characteristics]{A typical use case of gaze-tracking in portable devices and dataset characteristics}
\end{figure}

Figure \ref{fig:usecase} depicts a typical use case of portable computing devices (e.g. laptop, tablet) where a user is looking at a device screen while the device front camera is facing the person. These devices are predominantly used in landscape mode where the user is located within 50cm -- 95cm (20" -- 37") of comfortable reading range and could be tracked by the RGB camera. These RGB cameras are ubiquitous in all modern computing devices and we intend to utilize this minimal configuration to provide a computer mouse-like \textit{pointing device gesture} for eye gaze.

\section{Experimental Setup}
\label{sec:experimental_setup}
Deep Neural Networks (DNNs) have evolved into sophisticated learning automata through breakthrough achievements in the development of novel network architectures (e.g. Residual Networks, Capsule Networks), normalization and regularization techniques (e.g. Batch Normalization, Dropout), and various learning optimizers (e.g. SGD, Adam, Adagrad). These networks learn directly from the data and eliminate the need for hand-crafted features. When the input to the network is an image, Convolutional Neural Networks (CNNs) are the de facto approach because of their ability to identify complex spatial features -- crucial in making accurate predictions. In this work, we employ appearance-based gaze modeling using deep convolutional neural networks for constrained gaze applications.

\begin{figure}[]
  \centering
  \includegraphics[width=0.80\linewidth]{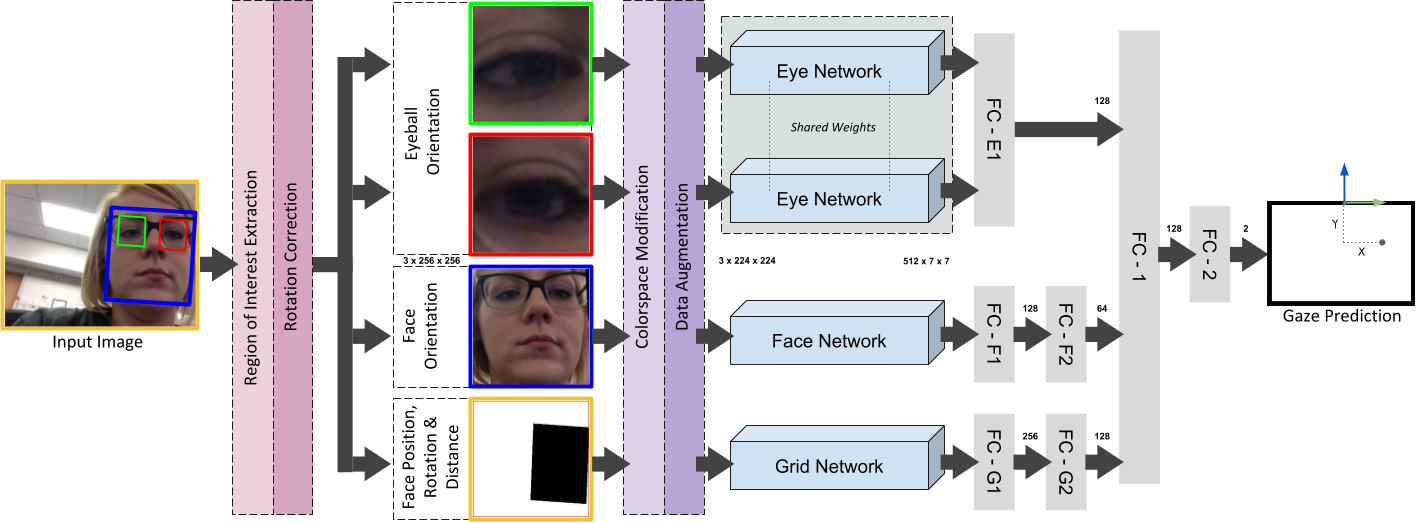}
  \captionsetup{justification=centering,margin=0cm}
  \caption{Enhanced iTracker Architecture}
  % label is given after the caption
  \label{fig:architecture_iTracker_Enhanced}
  \Description[Enhanced iTracker Architecture]{A 4-prong network architecture for Gaze-Tracking using both eyes, face and face-grid as input developed through enhancements over iTracker system}
\end{figure}

\textbf{Baseline:} For RGB-based constrained gaze-tracking, iTracker \cite{Krafka16} network architecture is the current state-of-the-art. Therefore, we reproduce it as an appropriate baseline for a series of experimental enhancements proposed in this paper. However, unlike iTracker we do not employ calibration or device-specific fine-tuning. The iTracker architecture takes the left eye, right eye, and face region crops from the original image as input along with a 25x25 binary face-grid indicating positions of all the face pixels in the original captured image. These input images are passed through Eye and Face sub-networks, which are based on AlexNet \cite{Krizhevsky12} as backbone architecture. Output of these region-specific sub-networks are processed through multiple fully-connected layers for the eventual gaze point coordinates estimation. Figure \ref{fig:architecture_iTracker_Enhanced} provides a high-level overview of the enhanced iTracker\footnotemark architecture. The experimental enhancements leading to its development are described in Section \ref{sec:experiments}.

\footnotetext{Please visit \url{https://github.com/MSREnable/GazeCapture} for further details and code related to this work.}

\textbf{Datasets:} To train a deep neural network that can be generalized to the real-world usage of portable computing devices, a large amount of data is required that covers various naturally occurring variations in the primary input signals (e.g. head pose, eye rotations) and operating settings (e.g. illumination, backgrounds). Many existing datasets for constrained gaze applications are either limited in their size \cite{Mora14}\cite{Sugano14}\cite{Zhang15}\cite{Huang17} or variations \cite{Weidenbacher07}\cite{McMurrough12}\cite{Smith13}. By far, the GazeCapture dataset \cite{Krafka16} -- acquired on smartphones and tablets -- is the largest publicly available dataset (1474 subjects and 2.4 million images) and for this reason, we use it for the experiments discussed in this paper. As face and eye regions are input to our model, in line with \cite{Krafka16}, we select only the frames that have valid face and eye detection. Detected regions of interest (i.e. face, left eye and right eye) are cropped and resized to 256px square image. Depending upon the use of data augmentation these images are resized to 224px square input images for the model either directly or through random cropping. To ensure a uniform data distribution during evaluation, the authors in \cite{Krafka16} exclude the subjects who did not look at the full set of points for validation and test data-splits. However to emulate a real-world scenario where a large-scale data collection may have various inconsistencies, we relaxed this constraint and restructured the dataset into a separate GazeCapture* dataset for a comparative study. We believe this would reveal the impact of data distribution on the accuracy and guide the research community during future dataset creation. Please refer to Figure \ref{fig:introduction} and Table \ref{tab:dataset} for a detailed distribution and characteristics of the datasets.

\textbf{Evaluation metrics and configuration:} We report the error in terms of mean Root Mean Square Error (a.k.a. Euclidean error) in centimeters from the ground-truth gaze point locations. We train a single model for both smartphone and tablet with the entire dataset and do not truncate the predictions based on the size of the device. For learning rate schedulers, learning rate was reduced from 1E-3 to 1E-4 (momentum 0.9 and decay 5E-4) over the 30 epochs of training in step-decay fashion. As described in Table \ref{tab:performance_evaluation}, Experiments 1-9 use a batch size of 128 and Experiments 10-14 a batch size of 100. 

\section{Experiments}
\label{sec:experiments}

\subsection{Regularization}
\label{ssec:regularization}

Convolutional Neural Networks stack a large number of convolutional layers together that makes them highly versatile in learning complex spatial patterns. The sheer volume the trainable parameters in them provide the capacity to represent any complex mathematical function but it also risks over-fitting - a phenomenon where the network starts learning various nuances in the data just to fit the training data too well. An effective technique to mitigate over-fitting is Regularization. The iTracker architecture uses Mean Image Subtraction (MIS) and Local Response Normalization (LRN) for image normalization and lateral inhibition. We replace them with Batch Normalization, which provides strong regularization capability through its control over the Internal Covariate Shift \cite{Ioffe15} and simplifying the learning dynamics between the layers to induce stable gradients. We also add Dropout layers (probability value 0.1) before each convolution and fully-connected layer in the network -- another powerful regularizer. Dropout layers randomly drop some internal connections, effectively creating several virtual sub-networks and prevent co-adaption.

\subsection{Data Augmentation}
\label{ssec:data_augmentation}
To achieve generalization a large amount of training data is required with variations that would be encountered during real-world use. When access to such large datasets is scarce, synthetically creating more data through small variations in original images provides a good alternative – a technique known as data augmentation. As a position augmentation, an original 256px square image was resized to 240px square image and then a 224px image was cropped at random. The brightness, contrast and saturation were randomly changed between [0.9, 1.1] and hue was varied between [-0.1, 0.1] for the color augmentation. During evaluation images were directly resized to 224px square image without color augmentation. The gaze point coordinates are not adjusted for the random cropping which adds some noise in the data and in an effect similar to Label Smoothening \cite{Muller19} prevents the model becoming overconfident. Horizontal image mirroring is also used as a positional augmentation in the last enhancement as listed in Table \ref{sec:results} but in this case, the gaze point coordinates are adjusted accordingly.

\subsection{Learning Rate Schedules}
\label{ssec:learning_rate_schedules}
Learning Rate is a crucial hyper-parameter which can yield faster convergence when tuned properly. Cyclic Learning Rate (CLR) \cite{Leslie17} practically eliminates this fine-tuning and yet achieves near optimal performance on a range of network architectures. Contrary to the traditional step-wise or exponentially decayed learning rate schedules, CLR periodically varies the learning rate within a band of values that prevent the model getting stuck in potential local minima or saddle points. In our experiments, we use CLR with a period of 8 epochs and vary the learning rate between 3E-3 and 5E-4. 

% Edited out
% Thus, providing a faster convergence at the cost of intermittent short-term divergences. This influence of saddle points during training is studied in detail by an independent work \cite{Dauphin14} where authors develop theoretical framework and tools to escape saddle points. It can also be argued that the optimal learning rate would very likely exist between the lower and the upper bound and thus, the model gets to use the near optimal learning rate amortized over time. 

\subsection{Residual Connections}
\label{ssec:residual_connections}

\begin{figure}[]
    \centering
    \begin{subfigure}[t]{0.245\linewidth}
        \includegraphics[trim=0.0cm 0.0cm 0.1cm 0.0cm,clip,width=1\linewidth]{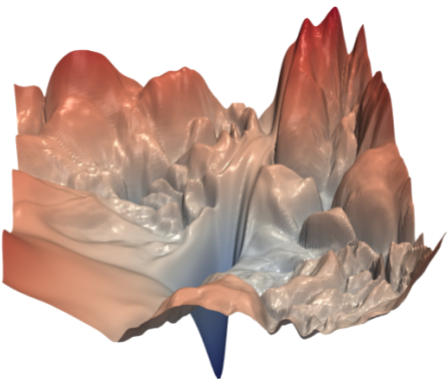}
        \caption{Without skip connections}\label{fig:loss_landscape_no_residual_connections}
    \end{subfigure}
    % leaving a space will take the subfigure to next line
    \begin{subfigure}[t]{0.245\linewidth}
        \includegraphics[trim=0.0cm 0.0cm 0.1cm 0.0cm,clip,width=1\linewidth]{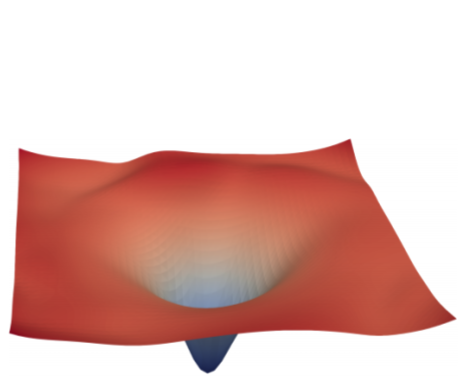}
        \caption{With skip connections}\label{fig:loss_landscape_residual_connections}
    \end{subfigure}
    \captionsetup{justification=centering,margin=0cm}
    \caption{The impact of residual "skip" connections on the loss landscape of ResNet-56. Retrieved from \cite{Li18}.}
    \label{fig:loss_landscapes}
    \Description[Loss-landscape smoothening through Residual Connections]{Impact of residual connections in smoothening the loss landscape.}
\end{figure}

% For a long time, deeper neural networks had higher training error than their shallow counterparts due to vanishing gradient problem. He et al \cite{He16}, first, demonstrated that adding some skip connections in the network eliminates this issue. These skip connections force the consecutive blocks in a neural network to learn residual functions instead of complex unreferenced functions and make it possible to train substantially deeper networks with increasingly higher accuracies. Recently, Li et al \cite{Li18} developed a ‘filter normalization’ method to visualize the loss landscapes and observed that these skip connections promote flat minimizers and prevent the transition from a nearly convex to otherwise a highly chaotic loss landscape (Figure \ref{fig:loss_landscapes}). We replaced the backbone architecture for the input branches of the model from AlexNet \cite{Krizhevsky12} to ResNet18 \cite{He16} architecture and utilized transfer learning over the pre-trained ResNet18 network as an initialization. Just like AlexNet backbone, we remove the classification layers (i.e. average pooling and fully connected layer).

He et al \cite{He16}, first, demonstrated that adding skip connections in deeper neural networks forces the consecutive blocks to learn residual functions instead of complex unreferenced functions and eliminates the vanishing gradient problem. This allowed training substantially deeper networks with increasingly higher accuracies. Recently, \textit{Filter Normalization} technique \cite{Li18} revealed that these residual connections promote flat minimizers and prevent the transition from a nearly convex to otherwise a highly chaotic loss landscape (Figure \ref{fig:loss_landscapes}). Therefore, we replaced the backbone architecture from AlexNet \cite{Krizhevsky12} to pre-trained ResNet18 \cite{He16} -- removing the classification layers (i.e. average pooling and fully connected layer) -- and utilized transfer learning.

\subsection{Color Transformation}
\label{ssec:color_transformation}
% For almost all the computer vision applications in deep learning, colored images are taken as input in the RGB color space. However, there are several other color spaces (e.g. HSV, YCbCr, LAB) which are used across other computer vision applications such as image compression, video transmission etc. It is natural to wonder whether there exists an optimal color space for a specific application. It can be argued that as most of these color spaces are linear transformation of RGB color space, a deep neural network should be able to learn an optimal color transform intrinsically as part of the training and therefore, the optimum performance should be independent of the employed color space. Nevertheless, studies \cite{Gowda19} have indicated that using an alternative or hybrid color space have actually improved the accuracy in certain class of applications. For example, it has been shown that in certain applications related to skin detection YCbCr color space yielded better accuracy than RGB color space \cite{Aibinu12}\cite{Doukim11}\cite{Shin02}\cite{AlMohair13}\cite{Shaik15}. YCbCr color space encodes color and intensity information separately making its channels highly independent compared to RGB and this quality could be attributed to the performance gain in such cases. As our model uses face and eye images as input, YCbCr color space becomes a natural candidate for a performance comparison against RGB color space. Also, we explore ImageNet-style mean and standard deviation based color normalization. 
Certain computer vision applications such as image compression and video transmission utilize non-RGB color spaces (e.g. HSV, YCbCr, LAB). Theoretically, a network should be able to perform various linear transformations itself to learn a better color space yet some studies \cite{Gowda19} have indicated that alternative or hybrid color spaces may significantly improve the accuracy. YCbCr colorspace, which encodes color and intensity information separately, is shown to perform better on certain applications related to skin detection \cite{Aibinu12}\cite{Doukim11}\cite{Shin02}\cite{AlMohair13}\cite{Shaik15}. Our model which has face and eye images as input becomes a natural candidate for performance evaluation with YCbCr color space. We also explore the influence of ImageNet-style mean and standard deviation based color normalization.

\subsection{Data Normalization}
\label{ssec:data_normalization}

Errors in the Region of Interest (RoI) extraction can have cascading effects in the gaze estimation pipeline. Further, it is crucial to normalize extracted face and eye images to the same size across the dataset to adjust for users' varying distance from the camera. The original detections in GazeCapture \cite{Krafka16} do not include above eyebrow and below lips regions. We compared it with two recent face detection libraries -- Apple face detection Circa 2019 and Dlib -- which resulted in a more useful, consistent and higher quality detection. We employed Dlib (version 19.19.0) for RoI cropping due to its superior performance and cross-platform compatibility. As depicted in Figure \ref{fig:plot_rotation_histogram}, there is a wide range of head rotation information in the captured images which needs to be extracted and encoded by the network intrinsically during the training for both face and eye images. Decoupling and efficiently passing this information to the network should increase the model capacity, while reduced input variance should make batch normalization more efficient. We utilized the OpenCV \textit{minAreaRect} method to fit a minimum area rectangle to the detected landmarks, estimate the head rotation angle, perform a homography based rotation correction, and extract face and eye crops. The rotation angle is encoded into the face-grid through a rotated grid which becomes the fourth input image to the model. The Dlib library detects some extreme head poses which were not detected originally and \textit{minAreaRect} method delivers tighter face regions increasing the overall RoI detection rate (Table \ref{tab:dataset}).

\section{Results \& Discussion}
\label{sec:results}

%% Data Distribution across the splits
\begin{table}[]
\captionsetup{justification=centering,margin=0cm}
\caption{Distribution of GazeCapture dataset over training and evaluation splits}
\begin{center}
\makebox[1.0\linewidth][c]{%centering table
\resizebox{1.0\linewidth}{!}{%resize table
\begin{tabular}{cccccccccc}
    \toprule
    \multirow{2}{*}{Dataset} & \multirow{2}{*}{Item} & \multicolumn{3}{c}{GazeCapture} & \multicolumn{3}{c}{GazeCapture*} & \multirow{2}{*}{Total} & \multirow{2}{*}{$\Delta$ Images}\\
    \cmidrule(lr){3-5}\cmidrule(lr){6-8}
     & & Train & Validation & Test & Train & Validation & Test & \\
    \midrule
    \multirow{2}{*}{Circa} & \#users & 1,271 & 50 & 150 & 1,018 & 299 & 154 & 1,471\\
    & \#images & 1,251,983 & 59,480 & 179,496 & 1,041,440 & 298,386 & 151,133 & 1,490,959\\
    \midrule
    \multirow{2}{*}{Dlib} & \#users & 1,272 & 50 & 150 & 1,020 & 298 & 154 & 1,472\\
        & \#images& 1,569,154& 72,869& 221,365 & 1,298,665 & 375,062 & 189,661 & 1,863,388 & +372,429\\
    \midrule
    \multirow{2}{*}{Dlib-RC} & \#users & 1,272 & 50 & 150 & 1,020 & 298 & 154 & 1,472\\
        & \#images & 1,605,280 & 74,194 & 226,228 & 1,328,934 & 382,414 & 194,354 & 1,905,702 & +414,743\\
    \midrule
    Data Splits & & 84\% & 4\% & 12\% & 70\% & 20\% & 10\% & 100\%\\
    \bottomrule
\end{tabular}
}}
\centering
\footnotesize
\item GazeCapture - Only the sessions with all datapoints are used in evaluation data-splits following \cite{Krafka16}.
\item GazeCapture* - Entire data is distributed across the splits in 70-20-10 ratio.
\end{center}
\label{tab:dataset}
\Description[GazeCapture data splits]{GazeCapture dataset is split across training, validation and test splits in 84\%, 4\% and 12\% respectively. For GazeCapture* the ratio is modified to 70\%, 20\% and 10\%. Using Dlib and rotation correction increases the valid face detection with about 415 thousand more images overall.}
\end{table}

% % Performance evaluation
\begin{table}[]
\captionsetup{justification=centering,margin=0cm}
\caption{Performance evaluation of various enhancement stages}
\begin{center}
\makebox[1.0\linewidth][c]{%centering table
\resizebox{1.0\linewidth}{!}{%resize table
\begin{tabular}{ccccccc}
    \toprule
    \multirow{2}{*}{Enhancement} & \multirow{2}{*}{S.N.} & \multirow{2}{*}{Details} & \multicolumn{2}{c}{GazeCapture} & \multicolumn{2}{c}{GazeCapture*}\\
    \cmidrule(lr){4-5}\cmidrule(lr){6-7}
    & & & Val Error & Test Error & Val Error & Test Error\\
    \midrule
    Baseline & 1 & AlexNet, RGB, [LRN, MIS], SGD & 2.0782 & 2.3239 & 2.3920 & 2.3819\\
    \midrule
    \multirow{2}{*}{Regularization} & 2 & AlexNet, RGB, [BN, DR, MIS], SGD & 6.7657 & 7.2043 & 6.9609 & 7.0001\\
    & 3 & AlexNet, RGB, [BN, DR], SGD & 2.0324 & 2.2010 & 2.3065 & 2.2821\\
    \midrule
    \multirow{2}{*}{Data Augmentation1} & 4 & AlexNet, RGB, BN, DR, SGD, [RandCrop] & 1.9851 & 2.1389 & 2.2579 & 2.2335\\
    & 5 & AlexNet, RGB, BN, DR, SGD, [RandCrop, Jitter] & 1.9558 & 2.1389 & 2.2695 & 2.2221\\
    \midrule
    \multirow{2}{*}{Learning Rate Scheduler} & 6 & AlexNet, RGB, BN, DR, [Adam], RandCrop, Jitter & 1.9796 & 2.1951 & 2.2892 & 2.2917\\
    & 7 & AlexNet, RGB, BN, DR, [CLR], RandCrop, Jitter & 1.9541 & 2.1057 & 2.2560 & 2.2132\\
    \midrule
    \multirow{2}{*}{Color Transformation} & 8 & AlexNet, [RGB, ISN], BN, DR, CLR, RandCrop, Jitter & 1.9417 & 2.1365 & 2.2144 & 2.1918\\
    & 9 & AlexNet, [YCbCr], BN, DR, CLR, RandCrop, Jitter & 1.9608 & 2.1148 & 2.2272 & 2.1874\\
    \midrule
    \multirow{2}{*}{Residual Connections} & 10 & [ResNet18-Frozen], YCbCr, BN, DR, CLR, RandCrop, Jitter & 3.1673 & 3.6192 & 3.4934 & 3.5731\\
              & 11 & [ResNet18], YCbCr, BN, DR, CLR, RandCrop, Jitter & 1.8366 & 1.9450 & 2.0609 & 2.0544\\
    \midrule
    \multirow{2}{*}{Data Normalization} & 12 & ResNet18, YCbCr, BN, DR, CLR, RandCrop, Jitter, [Dlib] & 1.8884	& 1.9463 & 2.0595 & 2.0862\\
              & 13 & ResNet18, YCbCr, BN, DR, CLR, RandCrop, Jitter, [Dlib, RC] & 1.8097 & 1.8397 & 1.9541 & 2.0245\\
    \midrule
    Data Augmentation2 & 14 & ResNet18, YCbCr, BN, DR, CLR, [RandCrop, Jitter, Mirror], Dlib, RC & 1.7612 & \textbf{1.8073} & 1.9183 & \textbf{1.9764}\\
  \bottomrule
\end{tabular}
}}
\centering
\footnotesize
\item LRN: Local Response Normalization, MIS: Mean Image Subtraction, SGD: Stochastic Gradient Descent, BN: Batch Normalization, DR: Dropout, CLR: Cyclic Learning Rate, RC: Rotation Correction
\end{center}
\label{tab:performance_evaluation}
\Description[Incremental performance improvement through enhancements]{Proposed enhancements in this work consistently improve the performance on both GazeCapture and GazeCapture* datasets.}
\end{table}

%%%%%%%%%%%% Figure: Error Plots %%%%%%%%%%
\begin{figure}[]
\centering
    \begin{subfigure}[t]{0.245\linewidth}
        \includegraphics[trim=0.5cm 0.2cm 1.5cm 0.9cm,clip,width=1\linewidth]{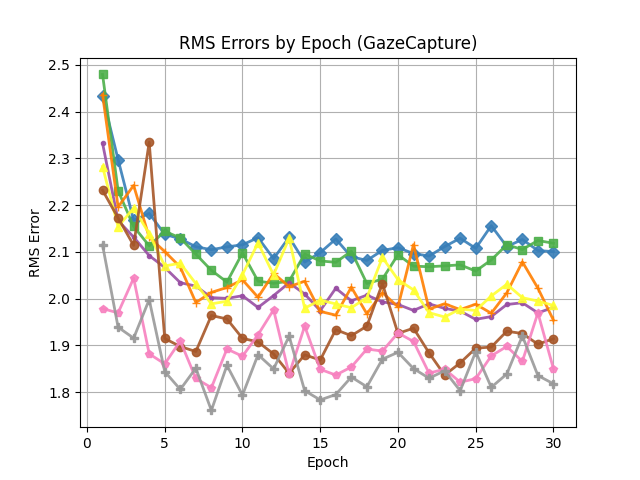}
        \caption{}\label{fig:plot_gc1_errorCurve}
    \end{subfigure}
    % leaving a space will take the subfigure to next line
    \begin{subfigure}[t]{0.245\linewidth}
        \includegraphics[trim=0.5cm 0.2cm 1.4cm 0.9cm,clip,width=1\linewidth]{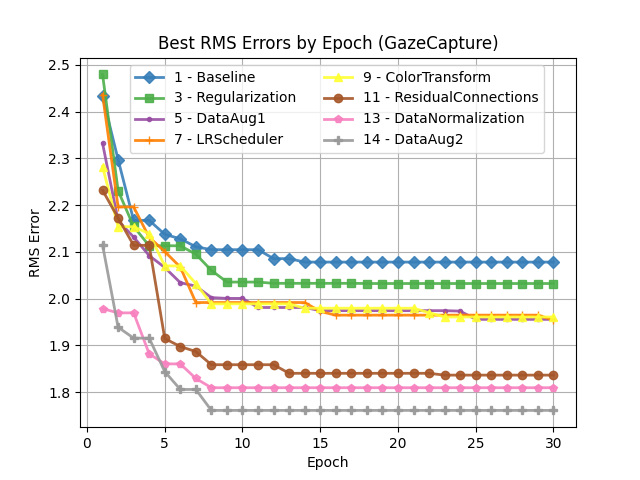}
        \caption{}\label{fig:plot_gc1_bestErrorCurve}
    \end{subfigure}
    \begin{subfigure}[t]{0.245\linewidth}
        \includegraphics[trim=0.5cm 0.2cm 1.4cm 0.9cm,clip,width=1\linewidth]{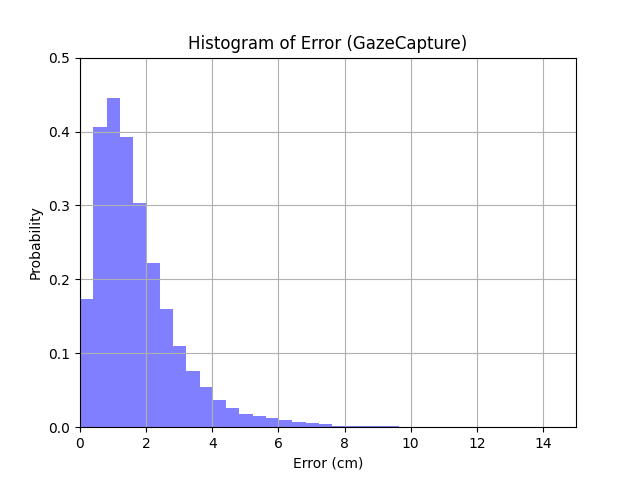}
        \caption{}\label{fig:error_histogram_gc1}
    \end{subfigure}
    \begin{subfigure}[t]{0.245\linewidth}
        \includegraphics[trim=0.5cm 0.2cm 1.4cm 0.9cm,clip,width=1\linewidth]{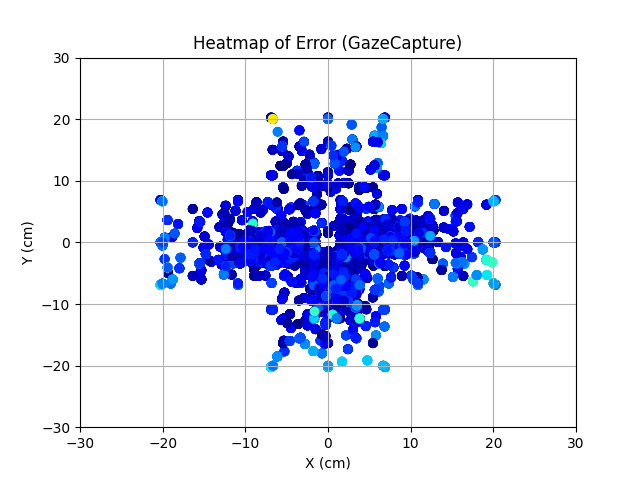}
        \caption{}\label{fig:error_heatmap_gc1}
    \end{subfigure}

    % leaving a space will take the subfigure to next line
    \begin{subfigure}[t]{0.245\linewidth}
        \includegraphics[trim=0.5cm 0.2cm 1.4cm 0.9cm,clip,width=1\linewidth]{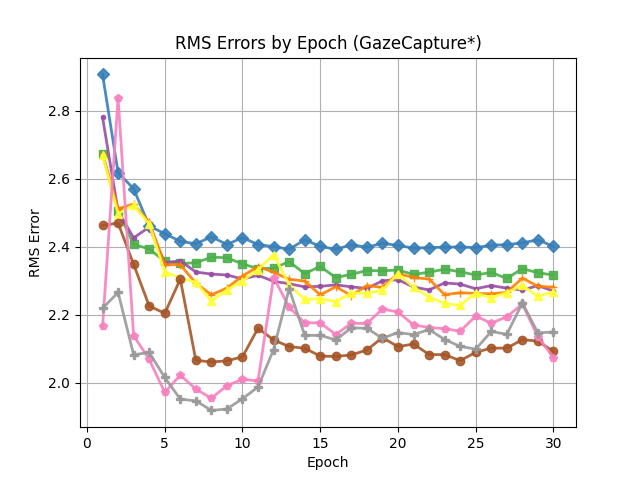}
        \caption{}\label{fig:plot_gc2_errorCurve}
    \end{subfigure}
    % leaving a space will take the subfigure to next line
    \begin{subfigure}[t]{0.245\linewidth}
        \includegraphics[trim=0.5cm 0.2cm 1.4cm 0.9cm,clip,width=1\linewidth]{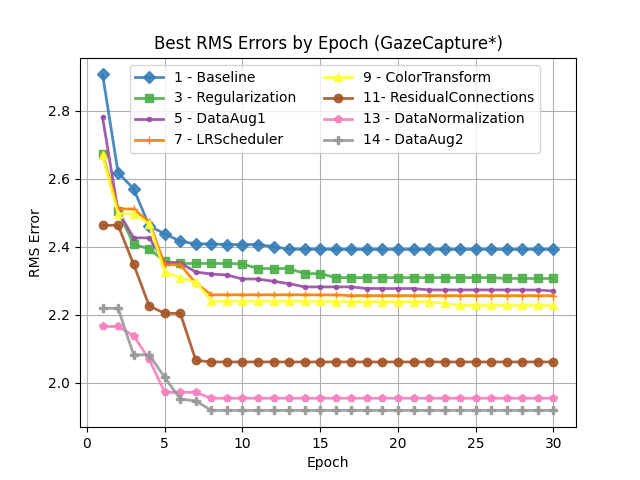}
        \caption{}\label{fig:plot_gc2_bestErrorCurve}
    \end{subfigure}
    \begin{subfigure}[t]{0.245\linewidth}
        \includegraphics[trim=0.5cm 0.2cm 1.4cm 0.9cm,clip,width=1\linewidth]{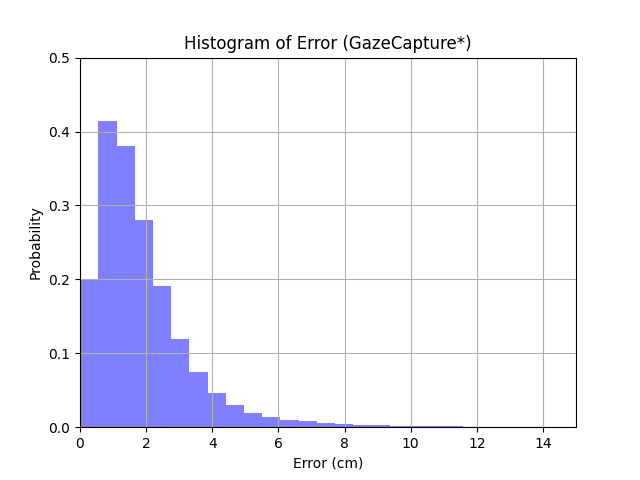}
        \caption{}\label{fig:error_histogram_gc2}
    \end{subfigure}
    % leaving a space will take the subfigure to next line
    \begin{subfigure}[t]{0.245\linewidth}
        \includegraphics[trim=0.5cm 0.2cm 1.4cm 0.9cm,clip,width=1\linewidth]{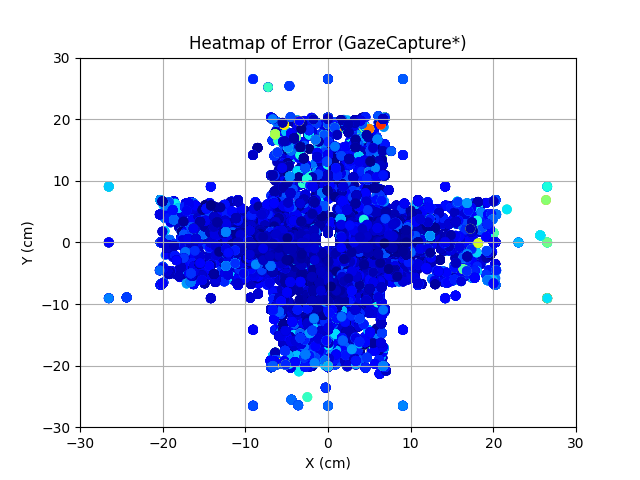}
        \caption{}\label{fig:error_heatmap_gc2}
    \end{subfigure}
    % leaving a space will take the subfigure to next line
  
    \begin{subfigure}[t]{0.33\linewidth}
        \includegraphics[trim=0.0cm 2.6cm 0.0cm 2.6cm,clip,width=1\linewidth]{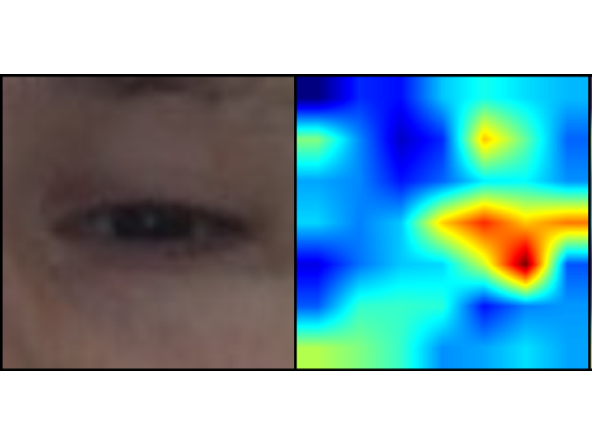}
        \caption{}\label{fig:gramCamPP_visualization}
    \end{subfigure}
  
  \captionsetup{justification=centering,margin=0cm}
  \caption{Error plots for (a)-(d) GazeCapture and (e)-(h) GazeCapture* respectively. (i) Grad-Cam++ visualization of an eye image}
  \label{fig:plot_errors}
  \Description[Error plots]{Error plots of various enhancement stages over GazeCapture and GazeCapture* datasets.}
\end{figure}

Table \ref{tab:performance_evaluation} lists the proposed experimental enhancements described in section \ref{sec:experiments} and their performances. Figure \ref{fig:plot_errors} depicts the corresponding mean-RMSError curves on the validation data-split over the 30 epochs of the training. Batch Normalization and Dropout for Data Regularization (Experiment 2-3) reduce the overall error, however, when combined with Mean Image Subtraction error is drastically increased. Batch Normalization normalizes the data over the batches to keep it well distributed for stable gradients encouraging higher learning rates. In contrast, Mean Image Subtraction normalizes every single image and changes the underlying data distribution. We believe, because of this Batch Normalization loses full control over the mean, variance and the gradients. Both random cropping and random color-jitter (Experiment 4-5) further improve the accuracy of the model. We do not see any accuracy improvement with Adam however Cyclic Learning Rate helped achieving lower error rates very quickly due to its periodic increment in learning rate before intermittent divergences (Experiment 6-7). It generally yields lower error rates at the end of the cycle. We observe mixed results with the use of ImageNet-style mean and standard deviation based image normalization and a minor performance gain through converting the input images into YCbCr color space (Experiment 8-9). Using the ResNet18 backbone provided the biggest performance boost due to underlying residual connections that make loss landscape smoother and pre-trained weights which provide a better weight initialization towards transfer learning. The impact of transfer learning becomes more evident when the pre-trained weights are not allowed to change which leads to performance degradation instead (Experiment 10-11). Data Normalization through Dlib face detection and rotation correction (Experiment 12-13) resulted into a significant improvement in performance which indicates a scope of future work in this direction to decouple various operating factors from the input data and pass it more efficiently. Horizontal Data Mirroring (Experiment 14) provides a fast method to virtually double the size of the training data and drop the error rate further.

It is important to understand what the deep learning model is learning to root out potential biases and other issues early in the development phase. For this, we utilized the Grad-Cam++ \cite{Chattopadhay18} technique to generate a heatmap of the model's internal gradient activities to encourage explainability. Figure \ref{fig:gramCamPP_visualization} is an example where we have an input eye image and corresponding activity heatmap on its right. As expected, we have a lot of activity around the eye region which indicates that the model cares a lot about the eye itself. But interestingly there is some activity around the eye-brow and the lower edge of the eye-lid which makes sense from a physiological perspective because those muscles are activated when we move our eye in certain directions. Trigonometric models that focus only at the pupil and the iris would not necessarily pick these features and therefore, this is where deep learning could exploit beyond the obvious in order to improve the accuracy.

\section{Conclusions}
\label{sec:conclusion}
In this paper, we propose an appearance-based deep learning pipeline for constrained gaze estimation. The technique utilizes ordinary RGB imagery captured on portable computing devices to regress the gaze point indicator making it hardware-agnostic. We reproduce the state-of-the-art iTracker architecture and benchmark several novel enhancements against it to significantly improve the accuracy. We demonstrate the versatility of residual connections and regularization in smoothening the loss landscape and maintaining stable gradients. We also show how normalizing the input through a consistent landmark detection and decoupling the rotation information could lead to better performance. Through incremental experiments and extensive evaluation, the proposed system achieved mean RMSError of 1.8073cm on the GazeCapture dataset without any calibration, fine-tuning or test-time data augmentation. We study the importance of a uniform data distribution. In its absence, we observe a consistent gap to the performance of the model that yielded 1.9764cm mean-RMSError on GapCapture*. We utilized the Grad-Cam++ technique to obtain visual explanations of the model's inter-workings which revealed the influence of the eye-center and muscular activities around it in the gaze point estimation. Gaze-tracking as accessibility technology has many roadblocks including lack of interoperability and non-existence of a diverse and large-scale dataset covering issues of facial occlusion, extreme head poses and various eye conditions. In the future, we intend to work in this direction and develop custom neural network architectures that can improve the performance even further.

%% The acknowledgments section is defined using the "acks" environment
%% (and NOT an unnumbered section). This ensures the proper
%% identification of the section in the article metadata, and the
%% consistent spelling of the heading.
\begin{acks}
% We would like to express our great appreciation to Team Gleason Foundation for their relentless efforts towards development of various accessibility technologies empowering people living with ALS around the world. Their partnership and feedback are a constant source of motivation for us. We would also like to extend our thanks to Krakfa, Khosla, Kellnhofer et al who contributed the iTracker architecture -- an inspiration for our current work -- and GazeCapture -- one of the biggest and most diverse publicly available datasets for gaze-tracking -- to the entire research community. 
We express our great appreciation to the Team Gleason Foundation for their relentless efforts towards development of various accessibility technologies, empowering people living with ALS around the world. Their partnership and feedback are a constant source of motivation for us. We would also like to extend our thanks to Krakfa, Khosla, Kellnhofer et al who contributed the iTracker architecture and GazeCapture dataset to the entire research community. Their dataset is by far one of the biggest and most diverse publicly available data for gaze-tracking and their work on the iTracker architecture is an inspiration for our current work.
\end{acks}

\bibliographystyle{unsrt}  
\bibliography{main}

\end{document}